\documentclass[a4paper,fleqn]{cas-dc}
\usepackage{natbib}
\usepackage{subfig}
\usepackage{hyperref}       % hyperlinks
\usepackage{url}            % simple URL typesetting
\usepackage{amsfonts}       % blackboard math symbols
\usepackage{nicefrac}       % compact symbols for 1/2, etc.
\usepackage{microtype}      % microtypography
\usepackage{xcolor}         % colors
\usepackage{amsmath}
\usepackage{amssymb}
\usepackage{graphicx}
\usepackage{tabularx}
\usepackage{tabulary}
\usepackage{multirow}
\usepackage{paralist}
\usepackage{enumitem}
\usepackage{booktabs}
\usepackage{caption}
\usepackage{bm}
\usepackage{amssymb}
\usepackage[switch]{lineno}
\usepackage[ruled,linesnumbered,lined,boxed,commentsnumbered]{algorithm2e}
\usepackage{threeparttable}

\newcommand{\tablestyle}[2]{\setlength{\tabcolsep}{#1}\renewcommand{\arraystretch}{#2}\centering\footnotesize}

\begin{document} %avoid word overflow
% \begin{sloppypar}
% \let\WriteBookmarks\relax
% \def\floatpagepagefraction{1}
% \def\textpagefraction{.001}
\shorttitle{ }
\shortauthors{ }

\title [mode = title]{Unifying UAV Cross-View Geo-Localization
via 3D Geometric Perception}
% {Unified Geometric Perception: Bridging the Gap Between Retrieval and Alignment for UAV Geo-localization}

% {3D-Aware Cross-View Alignment for Unified UAV Geo-localization}     

\address[1]{School of Electronic Information, Wuhan University, Wuhan 430072, China}
% \address[2]{School of Computer Science, Wuhan University, Wuhan 430072, China}
\address[2]{School of Artificial Intelligence, Wuhan University, Wuhan 430072, China}
\address[3]{Technology and Engineering Center for Space Utilization and the Key Laboratory of Space Utilization, Chinese Academy of Sciences, Beijing 100094, China}
\address[4]{School of Aeronautics and Astronautics, University of Chinese Academy of Sciences, Beijing 100049, China}

\cortext[cor1]{Corresponding author}

\author[1]{Haoyuan Li}

\author[1]{Wen Yang*}

\author[2]{Fang Xu}

\author[3]{Hong Tan}

\author[1]{Haijian Zhang}

\author[3,4]{Shengyang Li}

\author[2]{Gui-Song Xia}

\begin{abstract}
    Cross-view geo-localization for Unmanned Aerial Vehicles (UAVs) operating in GNSS-denied environments remains challenging due to the severe geometric discrepancy between oblique UAV imagery and orthogonal satellite maps. Most existing methods address this problem through a decoupled pipeline of place retrieval and pose estimation, implicitly treating perspective distortion as appearance noise rather than an explicit geometric transformation. In this work, we propose a geometry-aware UAV geo-localization framework that explicitly models the 3D scene geometry to unify coarse place recognition and fine-grained pose estimation within a single inference pipeline. Our approach reconstructs a local 3D scene from multi-view UAV image sequences using a Visual Geometry Grounded Transformer (VGGT), and renders a virtual Bird's-Eye View (BEV) representation that orthorectifies the UAV perspective to align with satellite imagery. This BEV serves as a geometric intermediary that enables robust cross-view retrieval and provides spatial priors for accurate 3 Degrees of Freedom (3-DoF) pose regression. To efficiently handle multiple location hypotheses, we introduce a Satellite-wise Attention Block that isolates the interaction between each satellite candidate and the reconstructed UAV scene, preventing inter-candidate interference while maintaining linear computational complexity. In addition, we release a recalibrated version of the University-1652 dataset with precise coordinate annotations and spatial overlap analysis, enabling rigorous evaluation of end-to-end localization accuracy. Extensive experiments on the refined University-1652 benchmark and SUES-200 demonstrate that our method significantly outperforms state-of-the-art baselines, achieving robust meter-level localization accuracy and improved generalization in complex urban environments.
    The data, model, and code will be released at \hyperlink{https://github.com/Collebt/Uni-CVGL}{https://github.com/Collebt/Uni-CVGL}.
\end{abstract}

\date{}
\maketitle
% introduction

\begin{figure}
    \centering
    \includegraphics[width=1\linewidth]{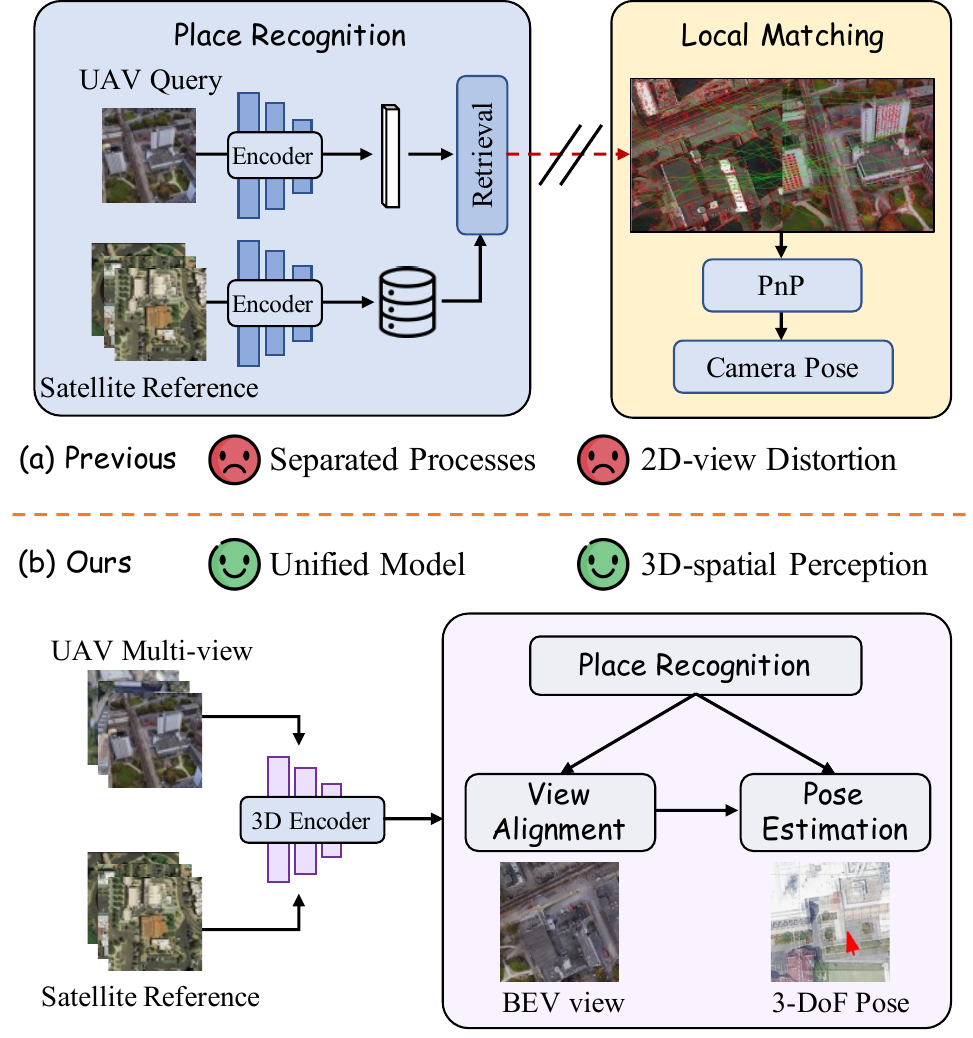}
    \caption{\textbf{Framework Comparison.} We show the comparison with existing two-stage UAV geo-localization methods and our unified UAV geo-localization framework.}
    \label{fig:compare_pipe}
\end{figure}

\section{Introduction}

Unmanned Aerial Vehicle (UAV) geo-localization serves as the cornerstone for autonomous navigation in GNSS-denied environments \citep{workman2015wide, zheng2020university, zhu2023sues}. A fundamental challenge in this task lies in cross-view matching between oblique UAV imagery and orthogonal satellite maps. Due to the drastic viewpoint discrepancy, vertical structures that are clearly visible from the UAV perspective are often collapsed into roofs or become completely invisible in satellite images, resulting in the well-known facade-to-roof ambiguity \citep{ye2024coarse}. This severe geometric inconsistency makes reliable cross-view correspondence extremely difficult and remains a long-standing bottleneck for UAV geo-localization systems. 

Most existing UAV cross-view geo-localization methods\citep{shetty2019uav, li2024learning} address this challenge through a decoupled pipeline, where coarse satellite retrieval, view alignment, and pose estimation are treated as separate stages. In such pipelines, retrieval is typically performed using an appearance-based 2D encoder to extract global descriptors for feature matching, while pose estimation is conducted as a subsequent image registration step, as illustrated in Figure \ref{fig:compare_pipe} (a). Although this design simplifies optimization, features optimized for global place recognition often lack precise geometric interpretability, while pose estimation requires spatially consistent and geometry-aware representations. As a result, errors accumulated during retrieval cannot be effectively corrected during pose estimation, leading to unstable localization performance under large viewpoint changes. 

Recent advances \citep{wang2025vggt, mildenhall2021nerf, kerbl20233d} in geometry-aware visual representations provide an opportunity to revisit this problem from a unified perspective. In particular, foundation models with explicit geometric grounding, such as Visual Geometry Grounded Transformers (VGGT), enable the extraction of features that are simultaneously discriminative at the scene level and informative at the geometric level. Nevertheless, existing methods often exploit such representations only partially, either for retrieval or for alignment, without fully integrating them into a single coherent localization framework.

These observations motivate a geometry-centric formulation of UAV geo-localization, where \textbf{retrieval, view alignment, and pose estimation are no longer treated as isolated components, but as tightly coupled sub-tasks operating on a shared geometric representation.} Rather than relying on multiple task-specific features or stage-wise processing, we argue that all three tasks should be jointly supported by a single geometry-aware feature space, ensuring spatial consistency throughout the localization process.

In this paper, we propose a unified geometry-aware UAV geo-localization framework that integrates satellite retrieval, view alignment, and precise pose estimation within a single model. As illustrated in Figure \ref{fig:compare_pipe} (b), our approach reconstructs a local 3D scene from UAV image sequences and renders a satellite-aligned bird's-eye-view (BEV) representation for pose estimation, both of which are derived from a single 3D feature representation. This unified feature serves as the common geometric foundation for all localization sub-tasks: it supports coarse satellite retrieval, guides cross-view alignment through BEV rendering, and enables fine-grained 3-DoF pose regression within a geometry-consistent inference pipeline. By eliminating task-specific feature fragmentation, our framework effectively mitigates the facade-to-roof ambiguity and enforces geometric consistency across all stages of localization.

Our contributions extend beyond architectural unification to the structural design of cross-view interaction under multiple location hypotheses. In large-scale UAV geo-localization, retrieval and alignment are inevitably performed against a gallery of satellite candidates. When all retrieved candidates are processed within a shared attention space, standard attention mechanisms allow feature interference across competing hypotheses, leading to degraded geometric consistency and unnecessary computational overhead. To address this issue, we introduce a Satellite-wise Attention Block, which enforces hypothesis-isolated interaction between the reconstructed UAV scene and each satellite candidate. Instead of jointly attending to all satellite features, our design conditions the cross-view interaction on individual satellite hypotheses, ensuring that each candidate is aligned independently within its own feature context. This structural isolation not only preserves the geometric integrity required for precise alignment and pose estimation, but also reduces the computational complexity of multi-candidate registration from quadratic to linear with respect to the number of satellite candidates.

Finally, to rigorously evaluate fine-grained localization performance under this unified setting, we release a recalibrated version of the University-1652 dataset with precise 3-DoF pose annotations and spatial overlap-aware evaluation protocols. Extensive experiments demonstrate that this geometry-first and hypothesis-isolated design significantly outperforms state-of-the-art baselines, achieving robust meter-level localization accuracy and strong zero-shot generalization across diverse environments.

The main contributions of this work are summarized as follows:
\begin{itemize}
\item We introduce a unified UAV geo-localization framework that jointly performs satellite retrieval, view alignment, and pose estimation within a single model, using a shared geometry-aware VGGT feature representation.
% \item We introduce a virtual BEV synthesis module that bridges the extreme viewpoint discrepancy between oblique UAV images and orthogonal satellite maps, facilitating robust cross-view feature matching.
\item We propose a Satellite-wise Attention Block that enables interference-free alignment between the reconstructed UAV scene and multiple satellite candidates with linear computational complexity, improving robustness in large-scale search scenarios.
\item We contribute a recalibrated version of the University-1652 dataset with precise coordinate annotations and spatial overlap-aware evaluation protocols, facilitating rigorous assessment of end-to-end UAV geo-localization performance. Our model achieves robust performance in meter-level UAV geo-localization, demonstrating superior robustness in challenging large-scale environments.
\end{itemize}

The remainder of this paper is organized as follows. Section~\ref{sec:related_work} reviews related work on cross-view geo-localization, fine-grained pose estimation, and end-to-end geo-localization. Section~\ref{sec:method} describes the proposed framework in detail, including an overview of the pipeline, the satellite-wise attention module, and the unified camera pose prediction process. Experimental results are presented in Section~\ref{sec:experiment}, where we evaluate the proposed approach on multiple datasets and provide a comprehensive analysis of its robustness and generalization ability. Further discussion is provided in Section~\ref{sec:discussion}. Finally, Section~\ref{sec:conclusion} concludes the paper.

% related work
\section{Related Work}
\label{sec:related_work}

This section reviews recent advancements pertinent to this study, categorizing the literature into three primary streams: UAV cross-view retrieval, fine-grained image matching, and end-to-end geo-localization architectures.

\subsection{UAV Cross-View Retrieval}

Cross-view image retrieval constitutes the coarse-level phase of geo-localization, tasked with identifying the most relevant satellite patch from a large-scale database using an oblique UAV query \citep{zheng2020university,zhu2023sues}. Early approaches \citep{workman2015wide, krizhevsky2017imagenet} were predominantly based on Convolutional Neural Networks (CNNs) trained with metric learning objectives. Foundational works \citep{liu2019lending} employed Siamese architectures with VGG or ResNet backbones, optimizing Contrastive or Triplet losses to learn view-invariant global descriptors \citep{simonyan2014very, he2016deep}. To address the spatial information loss inherent in global pooling, partition-based strategies, such as the Local Pattern Network (LPN) \citep{wang2021each}, introduced square-ring partitioning to enforce rotation-invariant spatial correspondence. While this strategy became a cornerstone for processing nadir imagery, it struggles with the severe projective distortions characteristic of oblique UAV views.

Recently, the paradigm has shifted toward Vision Transformers (ViTs) \citep{dosovitskiy2020image} and large-scale Foundation Models \citep{woo2023convnext, oquab2024dinov}. Architectures such as TransGeo \citep{zhu2022transgeo}, Sample4Geo \citep{Sample4Geo2023}, and FSRA \citep{Dai2022ATF} leverage self-attention mechanisms to capture long-range dependencies, implicitly modeling geometric layouts to achieve state-of-the-art retrieval performance. More recently, research has focused on bridging the ``domain gap'' via novel view synthesis and multimodal learning. For instance, the Panorama-to-BEV network \citep{ye2024cross}, Satellite-to-Street models \citep{toker2021coming}, and UAV-to-Satellite networks \citep{uav2022tian} explicitly synthesize ground or BEV representations to mitigate perspective discrepancies. Concurrently, multimodal approaches like MMGeo \citep{ji2025mmgeo} and CrossText2Loc \citep{ye2025cross} augment visual features with textual descriptions, enhancing retrieval robustness in repetitive environments through semantic reasoning.

\subsection{Fine-Grained UAV Image Matching}

While retrieval constrains the search space, fine-grained matching is essential for achieving meter-level precision. This task involves establishing dense pixel-wise correspondences between perspective UAV images and orthographic satellite maps. Conventional handcrafted descriptors, such as SIFT \citep{lowe2004distinctive}, lack robustness against the extreme viewpoint changes inherent in this domain. Consequently, the field has adopted deep feature matching paradigms. The combination of SuperPoint and SuperGlue \citep{detone2018superpoint, sarlin2020superglue, lindenberger2023lightglue} marked a significant advancement, utilizing graph neural networks to reject geometrically inconsistent outliers effectively.

The current state-of-the-art has evolved toward detector-free methods like LoFTR series \citep{sun2021loftr, wang2024efficient}, which perform dense or semi-dense matching on coarse feature grids using transformers. It demonstrates superior efficacy in low-texture regions where traditional keypoint detectors fail. However, geometric verification remains a bottleneck for these 2D-based approaches. Standard Perspective-n-Point (PnP) solvers \citep{zheng2013revisiting} typically rely on a ``planarity assumption'', treating the satellite map as a flat surface. This simplification introduces significant errors when matching facade points in UAV images to roof points in satellite maps, often leading to substantial pose drift in urban canyons \citep{dhaouadi2025ortholoc}.

\subsection{End-to-End Geo-Localization Architectures}

To scale visual geo-localization to large environments, early studies \citep{sarlin2019coarse, hughes2020deep, li2023jointly} adopted hierarchical pipelines that decouple global retrieval from local feature matching. While effective in practice, these designs rely on stage-wise optimization, which often leads to suboptimal representations for downstream pose estimation. To alleviate this limitation, more recent efforts have explored end-to-end architectures that integrate retrieval and matching into a single differentiable framework \citep{shetty2019uav, song2025unified}. In parallel, several works in the ground-to-aerial domain \citep{berton2022rethinking, trivigno2023divide} have investigated direct regression of meter-level locations from street-view imagery, aiming to bypass explicit matching altogether.

More recently, the emergence of 3D perception models has enabled a new class of UAV geo-localization approaches that incorporate explicit geometric reasoning \citep{li2025unsupervised, liu2025diffusionuavloc, ju2025video2bev}. These methods leverage 3D reconstruction paradigms, such as Neural Radiance Fields (NeRF) \citep{mildenhall2021nerf}, 3D Gaussian Splatting (3DGS) \citep{kerbl20233d}, and the Visual Geometry Grounded Transformer (VGGT) \citep{wang2025vggt}, marking a departure from purely implicit ``black-box'' regression. By explicitly reconstructing the underlying 3D scene, these approaches show potential to predict camera poses that are geometrically consistent with physical constraints, offering improved interpretability over heatmap-based or coordinate-regression models. Despite their promise, existing geometry-explicit methods face notable practical limitations. Many 3D reconstruction pipelines are computationally expensive, rely on slow Structure-from-Motion (SfM) initialization \citep{schonberger2016structure, pan2024glomap}, or are restricted to offline training-time augmentation. As a result, their deployment in large-scale or time-sensitive localization scenarios remains challenging. Consequently, the field still lacks a unified end-to-end framework that combines explicit geometric modeling with efficient and scalable inference.

From a broader perspective, current methods can be categorized into two paradigms: implicit models, which offer high computational efficiency but lack explicit geometric interpretability, and explicit reconstruction-based models, which provide accurate geometric reasoning at the cost of substantial computational overhead. Our work positions itself between these two extremes. By anchoring the localization pipeline on the VGGT, we retain the efficiency and generalization capability of implicit representations while enabling instant, explicit 3D scene prediction without SfM initialization. Building upon this representation, we synthesize a satellite-aligned BEV that orthorectifies the UAV perspective and enforces geometric consistency. This design yields a unified framework that jointly supports retrieval, alignment, and pose estimation, achieving a favorable balance between localization accuracy and inference efficiency.

\section{Methodology}
\label{sec:method}
To address the fundamental geometric inconsistency between oblique UAV imagery and orthogonal satellite maps, we propose a unified, geometry-aware UAV geo-localization framework that jointly performs place retrieval, view alignment, and absolute pose estimation within a single inference pipeline. Rather than treating these tasks as separate stages, our framework formulates them as coupled inference objectives supported by a shared geometric representation.

\begin{figure*}[ht]
	\centering
	\includegraphics[width=1\textwidth]{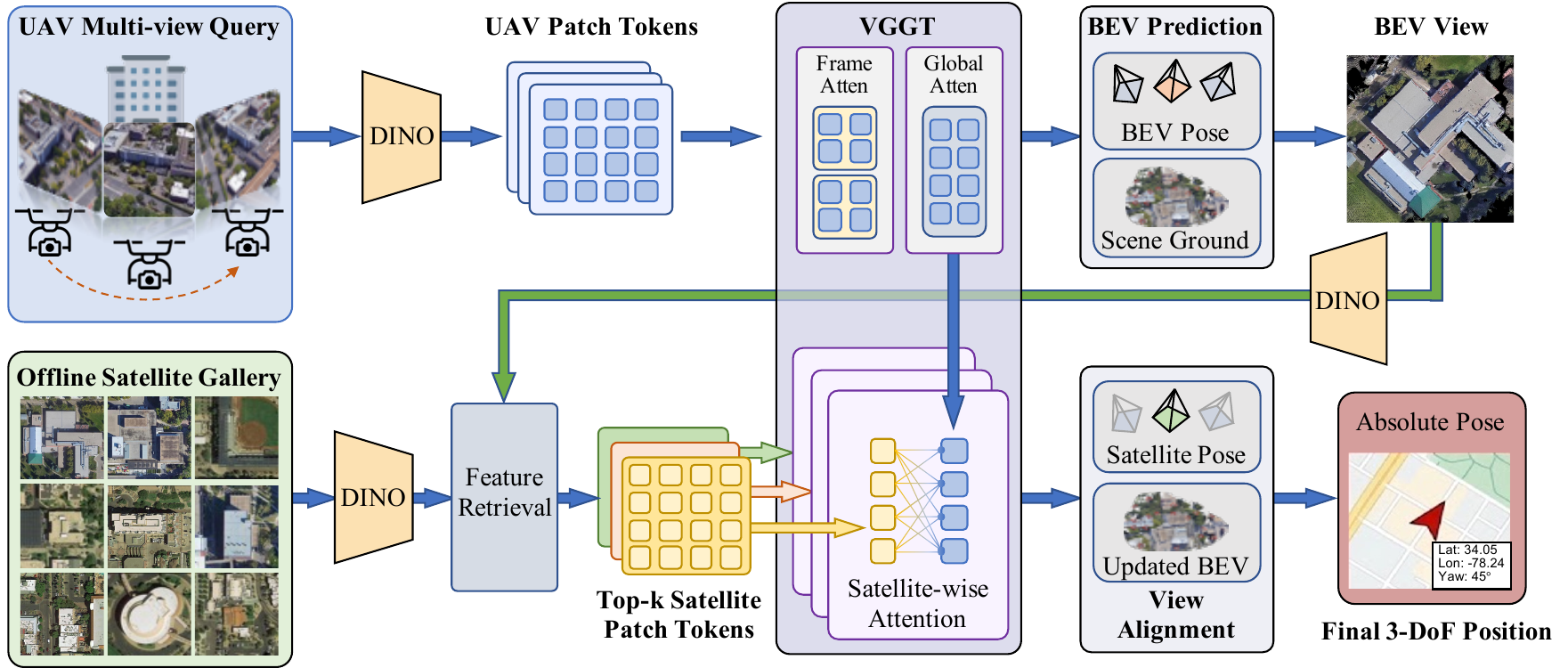}
	\caption{{\textbf{Overview of the unified UAV geo-localization pipeline.} Given a sequence of oblique UAV query images, the pipeline extracts the representation using VGGT to reconstruct a local 3D scene. Based on this shared representation, a satellite-aligned BEV is rendered to support coarse-level satellite retrieval. The retrieved satellite candidates are then independently aligned with the reconstructed 3D scene within the same feature space, enabling fine-grained regression of the absolute 3-DoF UAV pose (GPS coordinates and heading).}}
	\label{fig:pipeline}
\end{figure*}

\subsection{Unified UAV Geo-localization Framework}
As illustrated in Figure \ref{fig:pipeline}, a multi-view UAV query sequence and a satellite image gallery are processed to finally regress the absolute 3-DoF UAV pose (geographic position and heading). At the core of the framework lies a shared 3D perspective transformer, VGGT, which explicitly reconstructs the local scene geometry from the UAV inputs.
This reconstructed scene serves as a persistent geometric substrate throughout the pipeline, enabling retrieval, alignment, and pose estimation to be performed consistently within the same spatial context.

Specifically, VGGT simultaneously predicts a local 3D scene representation and synthesizes a satellite-aligned Bird's-Eye View (BEV). The BEV acts as a geometric bridge that orthorectifies the oblique UAV perspective, allowing direct interaction with satellite imagery while preserving spatial consistency. Crucially, all subsequent inference steps operate on this unified geometry-aware representation, without introducing decoupled task-specific features.

Based on this formulation, the proposed framework supports three tightly coupled inference objectives:
\textbf{(1) Unified Spatial Scene Representation}, which explicitly reconstructs the local 3D geometry and renders a satellite-aligned BEV;
\textbf{(2) Geometry-Aware Coarse Retrieval}, which performs place recognition directly in the BEV space to identify plausible satellite candidates;
and \textbf{(3) Fine-Grained Pose Estimation via View Alignment}, which integrates the retrieved satellite imagery with the reconstructed 3D scene to refine the absolute UAV pose. Together, these objectives form a single geometry-consistent inference process, enabling accurate and robust UAV geo-localization.

\textbf{Unified Spatial Scene Representation.}
To address the severe geometric deformations inherent between oblique UAV imagery and orthogonal satellite maps, we employ a spatial geometry model that projects both modalities into a unified canonical space. Initially, the UAV branch processes the $N$ input image sequence $\mathcal{I}^{(u)} = \{I^{(u)}_1, ..., I^{(u)}_N\}$ to predict an initial 3D scene representation. This procedure yields both the estimated extrinsic parameters of the UAV cameras and a dense point cloud of the underlying scene. Subsequently, the model generates a BEV estimation to explicitly simulate a satellite-aligned viewpoint from the reconstructed scene. This BEV representation serves as a spatial intermediary, transforming the UAV perspective into a global, orthogonal coordinate system compatible with satellite imagery. By unifying the geometric embedding space, we effectively mitigate the perspective distortions that typically degrade cross-view feature matching performance.

\textbf{Geometry-Aware Coarse Retrieval.}
In large-scale UAV geo-localization environments, efficient place recognition is a prerequisite for constraining the search space. Conventional methods typically extract global descriptors directly from UAV oblique images for matching against a satellite gallery. However, the extreme perspective disparity between oblique UAV photography and orthogonal satellite maps frequently results in feature misalignment and retrieval failure. To circumvent this limitation, we utilize the BEV representation to construct global features. Since the synthesized BEV shares the orthogonal projection characteristics of satellite imagery, it yields significantly more robust feature correspondences. We extract a global descriptor $\mathbf{g}^{(B)}$ from the predicted BEV image. Correspondingly, global descriptors are extracted from the satellite gallery, denoted as $\mathcal{G}_s = \{\mathbf{g}^{(s)}_{1},...,\mathbf{g}^{(s)}_{M} \}$. We formulate the place recognition task as a retrieval problem to identify the top-$k$ satellite candidates:

\begin{equation}
    \begin{array}{rl}
    C_k & = \{I^{(s)}_{1}, ..., I^{(s)}_{k}\} \\
        & = {\arg \max}_{\mathcal{G}_s} \{ d(\mathbf{g}^{(B)},\mathbf{g}^{(s)}_{1}),...,d(\mathbf{g}^{(B)}, \mathbf{g}^{(s)}_{M} )\},
    \end{array}
\end{equation}
where $C_k$ denotes the satellite candidates with top-$k$ highest score, and $d(\cdot, \cdot)$ denotes the distance between features. 

\textbf{Fine-Grained Pose Estimation.}
After identifying a set of plausible satellite candidates, the proposed framework refines the UAV location by performing geometry-consistent view alignment within the shared 3D representation. Rather than treating pose estimation as a separate post-retrieval step, alignment is formulated as the core geometric operation that jointly validates satellite candidates, updates the BEV view, and estimates precise camera poses.
Specifically, each retrieved satellite image is independently registered against the reconstructed 3D scene predicted by VGGT. Both UAV scene features and satellite features are embedded in the same feature space, enabling direct spatial correspondence without introducing task-specific representations. The model estimates relative camera poses that are consistent with the underlying scene geometry. The final absolute 3-DoF UAV pose is obtained by aggregating the alignment results across satellite candidates, favoring geometrically consistent hypotheses.

\subsection{Visual Patch Feature Extraction}
Visual patch features serve as the fundamental components for converting raw images into high-dimensional features for both 3D scene representation and global place recognition. The input image $I_i \in \mathbb{R}^{H \times W \times 3}$ is patchified into patches through a pre-trained DINOv2 backbone, yielding a set of patch tokens $\mathbf{t}^{(i)} \in \mathbb{R}^{L \times D}$, where $L$ is the number of patches and $D$ is the feature dimension. These tokens are then utilized in two distinct pathways:

\textbf{Geometry Scene Construction.} To capture the 3D structure of the scene, the patch tokens $\mathbf{t}^{(i)}$ of image $I_i$ are fed into the attention layers of the VGGT. Here, the tokens undergo frame-wise and global self-attention mechanisms to effectively model the spatial relationships and learn the underlying scene geometry. The output VGGT features are utilized for camera pose estimation and point cloud prediction.

\textbf{Global Feature Representation.} For coarse-level place recognition, we aggregate the patch tokens of BEV images and satellite images into a compact global descriptor $\mathbf{g}_i$. We employ Generalized Mean (GeM) pooling to suppress background noise and emphasize salient features:

\begin{equation}
    \mathbf{g}_i = \text{GeM}(\mathbf{t}^{(i)}) = \left( \frac{1}{L} \sum_{j=1}^{L} (\mathbf{t}^{(i)}_{j})^p \right)^{\frac{1}{p}},
\end{equation}
where $\mathbf{t}^{(i)}_{j}$ denotes the $j$-th token of image $I_i$ and $p=3$ is the pooling parameter. This yields a robust global representation suitable for efficient retrieval.

\subsection{Bird's Eye View Prediction}
Prior methods typically learn implicit semantic consistency between cross-view pairs directly from 2D images. However, these pairs often contain inherent view biases specific to the collection dataset, causing models to overfit and limiting generalization. We posit that effective cross-view geo-localization relies fundamentally on extracting spatial geometry perspectives from images. We hypothesize that geometry-centric models like VGGT hold significant potential for cross-view geo-localization by predicting a BEV representation that explicitly bridges the viewpoint gap. We then predict the BEV feature using the output point cloud from the VGGT prediction heads.

\textbf{BEV Camera Pose Prediction.} To synthesize a satellite-compatible view, we estimate a BEV camera, parameterized by extrinsics $[\mathbf{R}_\text{BEV} | \mathbf{T}_\text{BEV}]$, that aligns with the geometric ground plane of the reconstructed scene. 

First, we estimate the dominant ground plane $\pi_{ground}$ by analyzing the spatial distribution of the reconstructed 3D point cloud $\mathbf{p}$ using a plane fitting algorithm (e.g., RANSAC). We determine the plane parameters such that points $\mathbf{x}$ on the plane satisfy $\mathbf{n}^\top \mathbf{x} + d = 0$, where $\mathbf{n} \in \mathbb{R}^3$ denotes the unit normal vector and $d$ is the scalar offset. 

The rotation matrix $\mathbf{R}_\text{BEV}$ is constructed to align the virtual camera's optical axis with the scene's vertical axis (parallel to $\mathbf{n}$), ensuring the BEV camera plane is parallel to $\pi_{ground}$. To determine the translation vector $\mathbf{T}_\text{BEV}$, we position the virtual camera to cover the semantic center of the reconstructed scene. We calculate the centroid $\bar{\mathbf{p}}$ of the reconstructed 3D point cloud. The virtual camera center $\mathbf{C}_\text{BEV}$ is then placed above this anchor along the normal vector:
\begin{equation}
\mathbf{C}_\text{BEV} = \bar{\mathbf{p}} + \mathbf{n}.
\end{equation}

Finally, the translation vector is derived as $\mathbf{T}_\text{BEV} = -\mathbf{R}_\text{BEV} \mathbf{C}_\text{BEV}$. This formulation ensures that the synthesized BEV image effectively orthorectifies the scene, mitigating the perspective distortions that typically degrade cross-view feature matching performance.

\textbf{BEV Feature Aggregation.}
Once the virtual BEV camera pose is established, we aggregate the BEV features map for retrieval. Given the reconstructed scene point cloud $\mathbf{p} \in  \mathbb{R}^{N_p \times 3}$ and associated color features $\mathbf{c} \in  \mathbb{R}^{N_p \times 3}$, we first transform all points into the BEV camera coordinate system:
\begin{equation}
\mathbf{p}' = \mathbf{R}_\text{BEV} \mathbf{p} + \mathbf{T}_\text{BEV}.
\end{equation}

To generate a scale-invariant representation robust to varying scene densities, we employ an adaptive orthographic projection. The coordinates are normalized to the unit interval to fit within the viewport:
\begin{equation}
\mathbf{u} = \frac{\mathbf{p}' - \bar{\mathbf{p}}'}{2 \cdot S_{max}} + 0.5,
\end{equation}
where $S_{max}$ denotes the maximum spatial extent of point cloud, $\bar{\mathbf{p}}'$ denotes the centroid of transformed planar coordinates.

Finally, the normalized points $\mathbf{u}$ are discretized onto a downsampled pixel grid $\mathbb{1} \in \mathbb{R}^{H' \times W' \times 3}$. The pixel values at grid location $(u_x, u_y)$ are assigned the corresponding color values $\mathbf{c}$, resulting in the  BEV query feature used for satellite retrieval.

\subsection{Candidate Satellite Pose Estimation}
\begin{figure}
    \centering
    \includegraphics[width=0.95\linewidth]{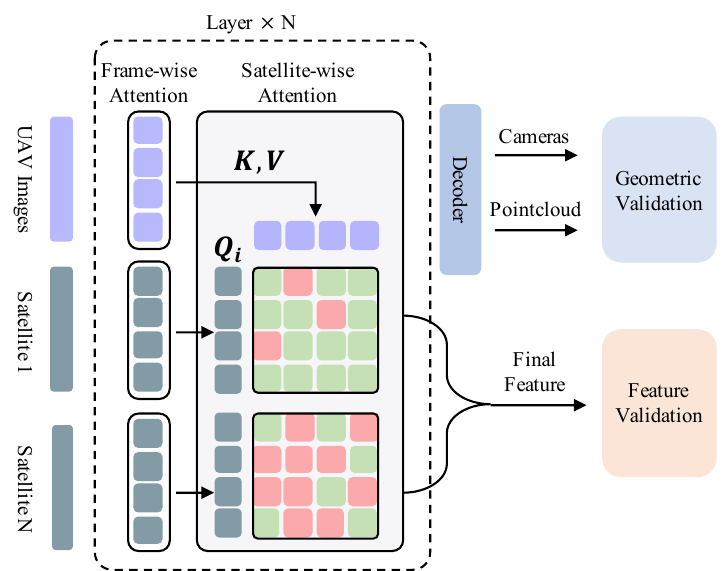}
    \caption{\textbf{Satellite-wise Attention Block.} The proposed satellite-wise attention block lets the satellite tokens attend to the UAV tokens independently, followed by the candidate validations.} 
    \label{fig:attn-block}
\end{figure}

To robustly anchor the local 3D reconstruction to the global satellite frame, we evaluate the geometric consistency of the retrieved satellite candidates relative to the UAV scene. This process involves effective candidate geometric prediction and selecting the optimal reference for final geo-localization, as illustrated in Figure \ref{fig:attn-block}.

\textbf{Satellite-wise Attention.} 
When passing the satellite candidates to the VGGT feature, this process necessitates an interaction between the satellite tokens and the UAV scene tokens on frame-wise attention and global attention. Let $\mathbf{t}^{(u)} \in \mathbb{R}^{N_u \times D}$ denote the set of UAV scene tokens derived from the VGGT reconstruction, and let $\mathbf{t}^{(s)} = \{\mathbf{t}^{(s)}_{i} \in \mathbb{R}^{N_{s} \times D}\}_{i=1}^K$ denote the sets of tokens for the $K$ retrieved satellite candidates. The conventional global attention is suboptimal for two primary reasons: 1) Allowing attention interactions between $\mathbf{t}^{(s)}_i $ and $\mathbf{t}^{(s)}_j (i\neq j)$ introduces noise and prevents the isolation of valid geometric cues.  2) The computational cost scales as $O((N_u + \sum N_{s})^2)$, which becomes prohibitive when processing dense satellite candidates.

To address these challenges, we propose a streamlined attention mechanism derived from the properties of the VGGT architecture. The VGGT employs an alternating structure of frame-wise attention and global attention, governed by residual learning dynamics:

\begin{equation}
    Z = \text{GlobalAttn}(\hat Z) + \hat Z,
\end{equation}
where $\hat Z$ represents the output of the preceding frame-wise attention layer.

We posit that the frame-wise attention is sufficient for aggregating the intrinsic context within a single satellite image (intra-view reasoning). Consequently, the subsequent global attention layer should be functionally decoupled to focus exclusively on inter-view geometric registration (alignment with the UAV scene). Decomposing the global attention update for the $i$-th satellite candidate reveals:

\begin{equation}
    \Delta Z^{(s)}_i  \! \!  \approx  \!\!  \! \! \! \underbrace{Attn(\text{Self})}_{\text{Redundant Update}}  \! \! \!  \! + \underbrace{Attn(\text{UAV}  \! \! \leftrightarrow  \! \! \text{Sat}_i)}_{\text{Geometric Alignment}} + \underbrace{Attn(\text{Sat}_i  \! \! \leftrightarrow  \! \! \text{Sat}_j)}_{\text{Irrelevant Noise}}.
\end{equation}

Since the intrinsic identity $\Delta Z^{(s)}_i$ is preserved via the residual connection, the $Attn(\text{Self})$ term provides marginal information gain. The 3-th term between satellites incurs noise and high computational cost. Therefore, we hypothesize that the geometric alignment can be approximated by exclusively calculating the cross-correlation between the satellite queries and the UAV scene.

Based on this derivation, we introduce the Satellite-wise Cross-Attention Block. Unlike standard global attention, we construct an asymmetric attention map where the satellite tokens serve as queries, while the UAV scene tokens exclusively serve as keys and values.

Formally, for the i-th satellite candidate $t^{(s)}_i$, we formulate the projection matrices as follows:
% \begin{equation}
%     \begin{cases}
%     \mathbf{Q}_i = \mathbf{W}_Q \mathbf{t}^{(s)}_i,\\
%     \mathbf{K}_i = \mathbf{W}_K \mathbf{t}^{(u)},\\
%     \mathbf{V}_i = \mathbf{W}_V \mathbf{t}^{(u)},
%     \end{cases}
% \end{equation}
\begin{equation}
    \mathbf{Q}_i = \mathbf{W}_Q \mathbf{t}^{(s)}_i, \quad
    \mathbf{K} = \mathbf{W}_K \mathbf{t}^{(u)}, \quad 
    \mathbf{V} = \mathbf{W}_V \mathbf{t}^{(u)}.
\end{equation}
Here, $\mathbf{K}$ and $\mathbf{V}$ are shared across all satellite candidates, computed solely from the UAV scene tokens. The update rule for the satellite tokens is then defined as:

\begin{equation}
    \text{Output}(\mathbf{t}^{(s)}_i) = \text{softmax}\left(\frac{\mathbf{Q}^{(s)}_{i} \mathbf{K}^\top }{\sqrt{d_k}}\right)V.
\end{equation}
The updated feature $Z^{(s)}_i$ is obtained by adding this cross-attention output to the residual stream $\hat Z^{(s)}_i$.

This design yields a substantial reduction in computational complexity. By pruning the satellite tokens from the Key/Value set, we transform the complexity from quadratic $O((N_u + \sum N_{s})^2)$ to linear with respect to the satellite candidates $O(N_u \cdot \sum N_s)$. This allows for the efficient parallel processing of multiple location hypotheses without cross-contamination, ensuring that the feature propagation remains theoretically sound via the residual path while maximizing inference speed.

\begin{figure}
    \centering
    \includegraphics[width=1\linewidth]{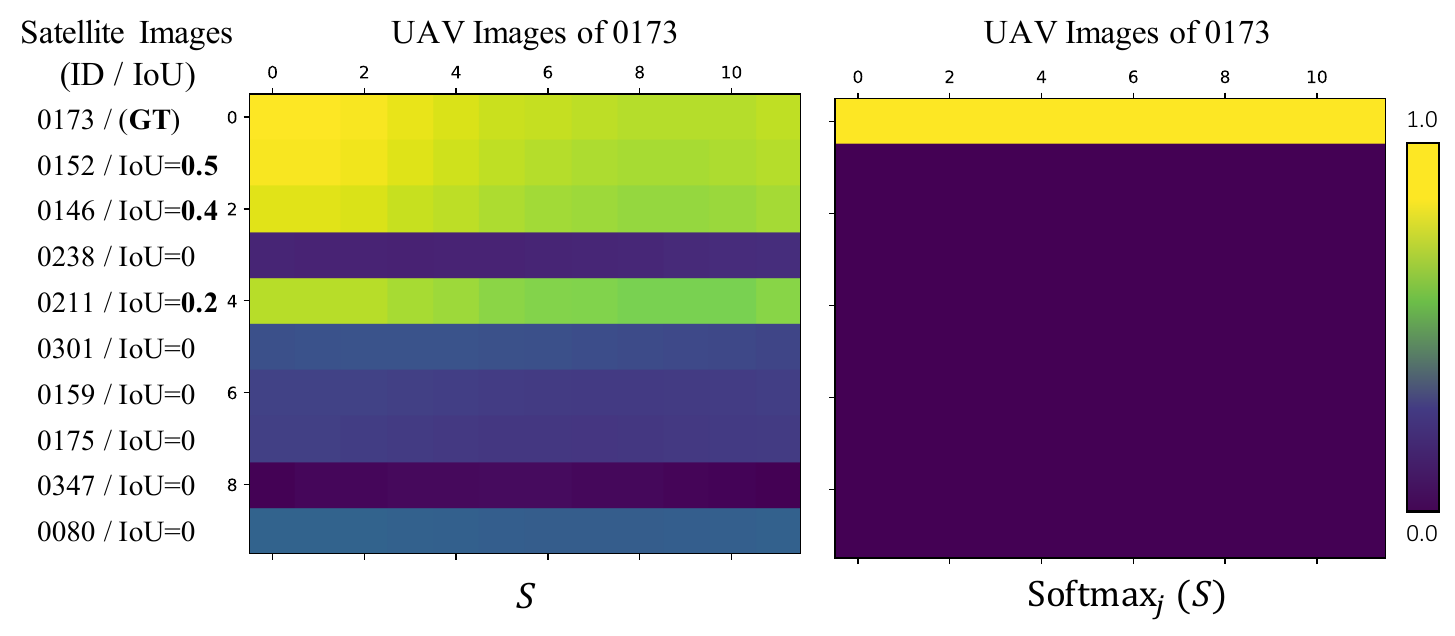}
    \caption{\textbf{Similarity Score of Features.}} 
    \label{fig:feat-sim}
\end{figure}

\textbf{Geometric Validation.} For each retrieved satellite candidate $I^{(s)}_i \in C_k$, we estimate its camera pose $[\mathbf{R}_\text{sat}$ | $\mathbf{T}_\text{sat}]$ in the world coordinate of the above UAV 3D scene. We project this camera pose into the initial BEV coordinate $[\mathbf{R}_\text{BEV}$ | $\mathbf{T}_\text{BEV}]$ for validation:
\begin{equation}
\mathbf{T}_i = (\mathbf{R}_\text{BEV})^\top (\mathbf{T}_\text{sat}^i - \mathbf{T}_\text{BEV}).
\end{equation}
This local translation $\mathbf{T}_i$ represents the offset of the satellite candidate from the reconstructed scene center. We then employ a dual-criteria validation strategy based on visual coverage and spatial proximity:

1) Visual Coverage: We calculate the number of valid 3D points $N_{valid}^i$ from the satellite view that exceed a dynamic confidence threshold $\tau_{conf}$. 

2) Spatial Proximity: We compute the $L_2$ distance offset between the estimated bev view and the satellite poses in the 2D horizontal plane $d_i = \|\mathbf{T}_i[0:2] - \mathbf{C}_\text{BEV}[0:2]\|_2^2$. 

From the set of valid candidates, we select the optimal satellite reference by maximizing a combined score that rewards high visual coverage and penalizes spatial distance:
\begin{equation}
S_i = \frac{N_{valid}^i}{\sum N_{valid}} - \frac{d_i}{\max(d)}.
\end{equation}
The candidate $I_s^*$ with the highest score $S^*$ is selected as the primary reference. Its pose $\mathbf{E}_{s}^*$ is then used to refine the BEV extrinsic $[\mathbf{R}_\text{BEV} | \mathbf{T}_\text{BEV}]$, and its corresponding high-confidence 3D points are merged into the rendering pipeline to enhance the final BEV visualization and pose estimation accuracy.

\textbf{Feature Similarity Validation.} 
Beyond explicit geometric constraints, we explore an implicit semantic verification strategy leveraging the rich feature representations learned by VGGT. We posit that the high-dimensional VGGT features encode sufficient geometric and semantic information to distinguish between true positive satellite matches and visually similar but geographically distinct distractors. To quantify this, we compute the feature similarity matrix between the retrieved satellite candidates and the UAV query sequence.

For the $i$-th satellite candidate and the $j$-th UAV image, the similarity score $S_{i,j}$ is computed as the mean similarity between their VGGT feature tokens:

\begin{equation}
S_{i,j} = \frac{1}{N M} \sum_{n=1}^{N} \sum_{m=1}^{M} \left( \mathbf{f}^{(i)}_{n} \mathbf{f}^{(j)\top}_{m} \right),
\end{equation}
where $\mathbf{f}^{(i)} \in \mathbb{R}^{N \times D}$ and $\mathbf{f}^{(j)} \in \mathbb{R}^{M \times D}$ denote the VGGT feature sets for the satellite and UAV images, respectively.

As visualized in Figure \ref{fig:feat-sim}, this similarity score $S$ (left) exhibits a strong correlation with spatial overlap. Satellite tiles that physically overlap with the UAV's field of view---even with a marginal Intersection-over-Union (IoU) as low as 0.2---yield significantly higher similarity scores compared to non-overlapping neighbors. This indicates that the VGGT features successfully capture shared geometric substructures robust to viewpoint changes. By applying a softmax operation over the candidate scores, the ground-truth tile (or the candidate with maximum overlap) consistently emerges with a dominant probability score (right). This peaked distribution provides a highly reliable signal for positive sample selection, effectively filtering out geometric outliers that lack semantic and geometric consistency.

\subsection{Absolute Pose Prediction}

The geometry-aware model outputs relative pose estimates for both the UAV cameras and the retrieved satellite images in a shared, scale-ambiguous local coordinate system. To recover the absolute geo-coordinates (GPS and heading), we align the local reconstruction with the real-world scale and the global coordinate frame defined by the satellite imagery.

\textbf{Spatial Scale Estimation.} We first determine the metric scale factor $\alpha$ relating the local reconstruction units to real-world meters. We extract high-confidence 3D points $\mathbf{p}_\text{sat}$ corresponding to the satellite view from the model's predictions. The scale is computed by comparing the spatial range of these points against the known physical resolution of the satellite image:

\begin{equation}
\alpha = \frac{\sqrt{(W \cdot \delta)^2 + (H \cdot \delta)^2}}{|\max(\mathbf{p}_\text{sat}) - \min(\mathbf{p}_\text{sat})|_2},
\end{equation}
where $W, H$ are the satellite image dimensions in pixels, and $\delta$ is the ground resolution (meters/pixel) calibrated from the database.

\textbf{UAV Coordinate Transformation.} Let $[\mathbf{R}_\text{UAV} | \mathbf{T}_\text{UAV}]$ and $[\mathbf{R}_\text{sat} | \mathbf{T}_\text{sat}]$ denote the estimated camera-to-world extrinsics for a UAV frame and the satellite reference image, respectively. We transform the UAV position into the satellite-centered frame:
\begin{equation}
\mathbf{T}_{rel} = \alpha \cdot \mathbf{R}_\text{sat}^\top (\mathbf{T}_\text{UAV} - \mathbf{T}_\text{sat}).
\end{equation}

\textbf{Global Geo-Registration.} Using the known GPS origin $(\text{lat}_0, \text{lon}_0)$ of the satellite image, we first convert this origin to meter-level UTM coordinate and calculate the global location of the UAV poses. We then convert the relative translation $\mathbf{T}_{rel} = [\Delta E, \Delta N, \Delta U]^\top$ into global UTM coordinates $\mathbf{T}_{abs} = (E_0 + \Delta E, N_0 + \Delta N)$. The absolute UAV heading $\theta_{abs}$ is derived by projecting the local camera z-axis (view direction) onto the horizontal plane of the satellite frame:
\begin{equation}
    \begin{array}{ccc}
        \mathbf{z}  &=& \mathbf{R}_\text{sat}^\top (\mathbf{R}_\text{UAV}  \mathbf{\hat z}) \\
        \theta_{abs}&=& \text{arctan}(\frac{\mathbf{z}_{y}}{\mathbf{z}_{x}}), \\
    \end{array}
\end{equation}
where $\mathbf{\hat z} = [0,0,1]^\top$ represents the optical axis in the camera. This process yields the final absolute 3-DoF pose $[\mathbf{T}_{abs}, \theta_{abs}]$ for each UAV query frame.

\section{Experiments}
\label{sec:experiment}

\subsection{University-Pose Geo-localization Benchmark}
To rigorously evaluate the fine-grained 3-DoF pose estimation capabilities of our framework, we extended the standard University-1652 dataset \citep{zheng2020university}. While the original dataset provides image-level correspondences, it lacks the precise absolute camera coordinates required for meter-level geo-localization evaluation. To address this, we recalibrated the dataset as \textbf{University-Pose} with 3-DoF UAV pose, by extracting and transforming the raw flight trajectory data from the Google Earth annotation files provided with the dataset. We derive the absolute camera coordinates from the target control parameters: longitude $\lambda$, latitude $\phi$, and heading $\theta$.

\begin{figure}
    \centering
    \includegraphics[width=1\linewidth]{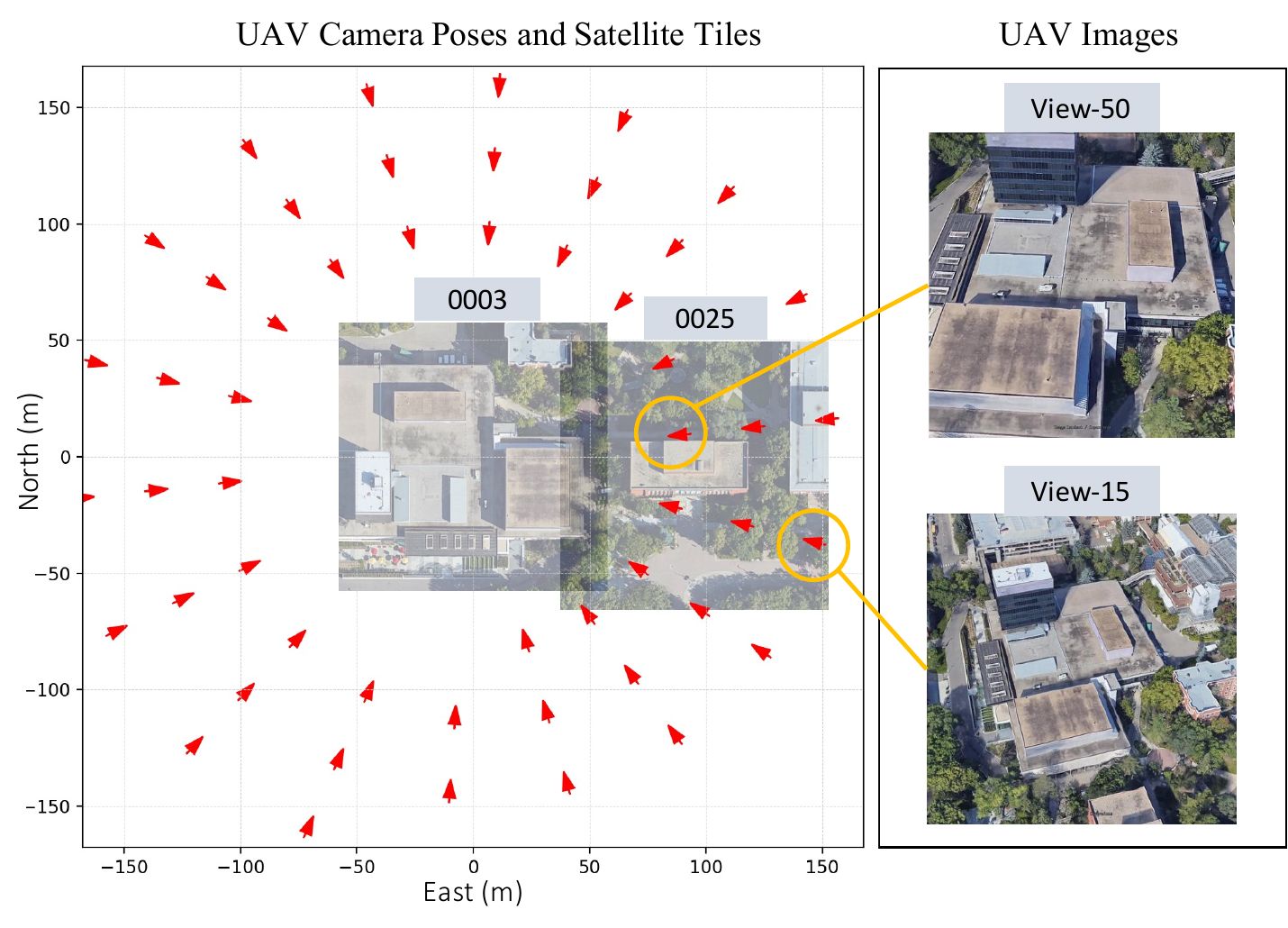}
    \caption{\textbf{Geometric Relationship between UAV Camera and Satellite Tiles.} Visualization of the University-Pose dataset geometry. The red arrow represents the 3-DoF UAV pose. Notably, the dataset contains spatially overlapped satellite tiles, providing dense candidates that facilitate robust pose estimation.} 
    \label{fig:pose-vis}
\end{figure}

The first critical enhancement in our recalibration is the explicit modeling of spatial overlaps in the satellite database. As illustrated in Figure \ref{fig:pose-vis}, the satellite tiles in the gallery are not disjoint but exhibit significant spatial overlap. This overlap is advantageous for fine-grained localization: when a UAV query retrieves multiple overlapping satellite candidates, the redundancy allows for multi-view geometric verification, significantly improving pose estimation robustness compared to single-tile matching. Figure \ref{fig:pose-vis} visualizes this geometric relationship, showing the precise alignment between the 3-DoF UAV camera poses and the corresponding satellite tile center, situated within a dense grid of overlapping satellite imagery.

\begin{figure}
    \centering
    \includegraphics[width=1.0\linewidth]{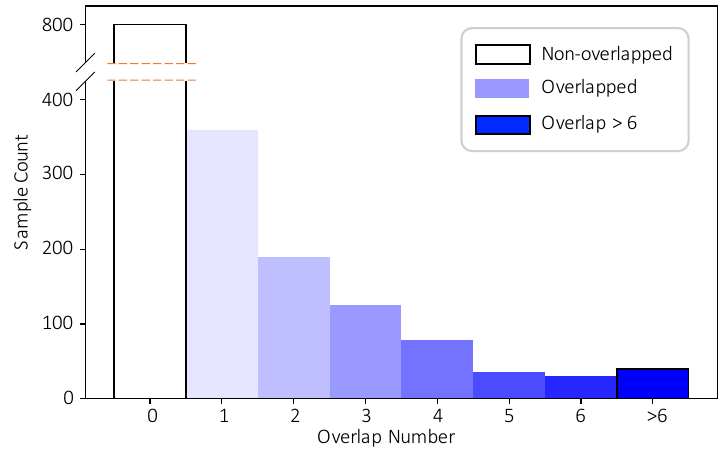}
    \caption{\textbf{Overlapped Satellite Images Statistic.}} 
    \label{fig:sat-overlap}
\end{figure}

We further quantify the spatial redundancy within the University-Pose dataset by analyzing the overlap distribution of satellite tiles. As depicted in Figure \ref{fig:sat-overlap}, we visualize the statistical distribution of the number of overlapping satellite tiles for every scene. Our analysis reveals that approximately 50\% of the scenes are covered by more than one satellite tile. This implies that retrieved neighboring tiles---often classified as false positives in strict binary evaluation---contain substantial visual overlap with the query and can provide helpful geometric information for subsequent fine-grained pose estimation. This observation motivates our proposal of a refined retrieval metric (see Section \ref{sec:metrics}) that considers these overlapped ``semi-positive'' samples, rather than relying solely on binary target identification.

\subsection{Evaluation Metrics}
\label{sec:metrics}

To comprehensively assess the performance of our unified framework, we adopt a dual-metric evaluation protocol covering both the coarse-level retrieval stage and the fine-grained absolute pose estimation stage. 

\textbf{Recall@K (R@K):} For the coarse-level place recognition task, we employ standard information retrieval metrics to evaluate the quality of the top-ranked satellite candidates. This metric measures the percentage of query images for which the ground-truth satellite match appears within the top-$K$ retrieved candidates. We report R@1, R@5, and R@10. R@1 is particularly critical as it reflects the model's ability to identify the correct location immediately without further verification.

\textbf{Average Precision (AP):} To account for the ranking quality of all relevant candidates, we compute the Average Precision (AP), which is defined as the area under the precision-recall curve. The mean Average Precision (mAP) is reported as the average AP across all query samples.

\textbf{IoU-based Recall (R${_\text{IoU}}$@K):} The standard Recall@K metric relies on a strict binary identification of the exact ground-truth image ID, often penalizing valid retrievals in datasets with spatially overlapping satellite tiles. In practice, neighboring tiles with significant overlap share substantial visual content with the UAV query, rendering them viable candidates for subsequent fine-grained pose estimation. To rigorously evaluate this, we define R${_\text{IoU}}$@K, which considers a retrieved candidate as a positive match if its spatial IoU with the ground-truth tile exceeds a predefined threshold $\tau$. This metric validates the retrieval of any geometrically usable reference image, offering a more robust assessment of place recognition performance in dense grids.

\textbf{Global Position Error (GPE):} For the fine-grained localization task, we evaluate the accuracy of the estimated 3-DoF absolute pose (latitude, longitude, and heading). The GPE quantifies the planar distance in meters between the predicted location $\mathbf{T}_{pred} = (\lambda_{pred}, \phi_{pred})$ and the ground truth location $\mathbf{T}_{gt} = (\lambda_{gt}, \phi_{gt})$ in the Universal Transverse Mercator (UTM) grid system:

\begin{equation}
    E_{pos} = \sqrt{(\lambda_{pred} - \lambda_{gt})^2 + (\phi_{pred} - \phi_{gt})^2}.
\end{equation}

\textbf{Heading Error (HE):} The heading error $\Delta \theta$ measures the absolute angular deviation between the predicted orientation $\theta_{pred}$ and the ground truth heading $\theta_{gt}$:

\begin{equation}
\Delta \theta = \min(| \theta_{pred} - \theta_{gt} |, \quad 360^{\circ} - | \theta_{pred} - \theta_{gt} |).
\end{equation}

\textbf{Meter-level Success Rate (SR@X):} To provide a strict measure of usability for autonomous navigation, we report the Meter-level Success Rate (MSR@X), defined as the percentage of test queries where the positioning error falls within a specific threshold $X$ (e.g., 3 meters, 5 meters, 10 meters). A query is considered successfully localized if and only if $E_{pos} \leq X$.

\begin{equation}
\text{MSR}@X = \frac{1}{N} \sum_{i=1}^{N} \mathbb{1}(E_{pos}^{(i)} \leq X),
\end{equation}
where $N$ is the total number of queries and $\mathbb{1}(\cdot)$ is the indicator function.

\subsection{Experimental Settings}
\label{implementationDetails}
\textbf{UAV Geo-localization Datasets.} We conduct comprehensive evaluations of our method on two prominent UAV-to-satellite retrieval datasets: University-Pose and SUES-200 \citep{zhu2023sues}. These benchmarks are renowned for the significant challenges imposed by the perspective discrepancy between oblique UAV imagery and orthogonal satellite maps. For the UAV query input, we randomly sample a continuous sequence of $N$ views from each scene, while the satellite gallery follows the standard test set protocols.

\textbf{Implementation Details.} Our framework is implemented using the PyTorch \citep{PyTorch_2019_NIPS} deep learning framework. We employ the pre-trained DINOv2 \citep{oquab2024dinov} network as the backbone for robust feature extraction. The 3D perspective modeling is handled by the VGGT. During inference, the system processes continuous sequences of UAV frames as input. Ground plane estimation is performed using the RANSAC algorithm with an inlier threshold of 0.1. The number of retrieved satellite candidates is set to $K=10$.

\subsection{Place Recognition Results}
\renewcommand{\arraystretch}{1.3}
\begin{table*}[]
    \caption{\textbf{Retrieval Result on University-1652 and SUES-200.}}
    \begin{tabular}{c|ccccccc|cc}
        \toprule
        \multirow{2}{*}{Methods}  & \multicolumn{7}{c|}{University-1652}  & \multicolumn{2}{c}{SUES-200 (150m)}\\
    \cline{2-10}
    & R@1 $\uparrow$     & R@5 $\uparrow$     & R@10  $\uparrow$  & AP  $\uparrow$     &  R$_\text{IoU}$@1   $\uparrow$  &  R$_\text{IoU}$@5 $\uparrow$ & Time(s) $\downarrow$ & R@1   $\uparrow$         & AP  $\uparrow$      \\
    \cline{1-10}
    University \citep{zheng2020university} & 69.3    & 86.7   & \textbf{91.2}   & 73.1         & -       & -   & \textbf{0.8}  & 67.8    & 72.7         \\
    AnyLoc \citep{keetha2023anyloc}        & 36.2    & 59.1   & 57.3   & 36.2         & 37.2       &60.2  & 1.9  & 30.0     & 42.7         \\
    SegVLAD \citep{garg2024revisit}        & 56.8    & 68.1   & 76.9   & 56.9         & 60.2       &69.3     & 16.1  & 56.0      & 59.1        \\
    UCVGL \citep{li2025unsupervised}       & 54.5    & 63.2   & 77.9   & 66.2         & 57.1       &74.1     & 62.3 & 56.0      & 60.5          \\
    \rowcolor{gray!20}Ours  (Initial BEV)  & 58.8    & 78.7   & 84.2   & 63.2         & 63.2       &79.2 & 3.5  & 56.5      & 61.6   \\
    \rowcolor{gray!20}Ours  (Refined BEV)  &\textbf{79.0}     & \textbf{88.0}         & 89.3   & \textbf{81.1}        & \textbf{83.9}       & \textbf{88.4}         & 9.6   & \textbf{68.0}  & \textbf{74.5}  \\
    \bottomrule
    \end{tabular}
    \label{tab:benchmark}
\end{table*}

\renewcommand{\arraystretch}{1.3}
\begin{table}
    % \tablestyle{10pt}{1.2}
    \caption{\textbf{IoU-based Metric on Retrieval Results.}}
    \begin{tabular}{c|ccc}
        \toprule
        IoU Threshold $\tau$  & R$_\text{IoU}@1  $$\uparrow$     & R$_\text{IoU}@5 $$ \uparrow$    & R$_\text{IoU}@10 $$\uparrow$ \\
        \cline{1-4}
        0.1	    &84.9	&89.2	&89.9 \\
        0.2	    &83.9	&88.4	&89.4 \\
        0.5	    &80.6	&88.3	&89.4 \\
        0.8	    &79.9	&88.2	&89.3 \\
        1.0	    &79.0	&88.2	&89.3 \\
        \bottomrule
    \end{tabular}
    \label{tab:iou_sensitivity}
\end{table}
\textbf{Retrieval Results.}
We present a comprehensive evaluation of our unified framework against existing state-of-the-art (SOTA) general methods on the University-1652 \citep{zheng2020university} and SUES-200 \citep{zhu2023sues} benchmarks, as summarized in Table \ref{tab:benchmark}. To ensure a fair comparison, for the task-specific method \citep{zheng2020university}, we report its multi-view performance using the m-query setting from \citep{zheng2020university, zhu2023sues}. For the zero-shot methods \citep{keetha2023anyloc, garg2024revisit}, we implement these single-frame methods by processing the same UAV sequence frames and averaging the output features as a global descriptor, isolating the effect of frame quantity. 

The results demonstrate that our method establishes a new standard for UAV geo-localization. On University-1652, characterized by complex urban geometries, our approach not only significantly outperforms the zero-shot baselines but also achieves superior Recall@1 and mAP scores compared to the supervised University-1652 baseline (m-query). Similarly, on the SUES-200 dataset, particularly at the challenging 150m altitude, our framework exhibits exceptional robustness to extreme viewpoint variations where traditional 2D matching methods typically fail. Crucially, this high precision does not come at the cost of speed; thanks to the linear complexity of our Satellite-wise Attention mechanism, our pipeline maintains high inference efficiency compared to computationally intensive iterative matching baselines.
Furthermore, a key driver of this performance is the framework's iterative refinement capability. We report results for two stages: the \textbf{Initial BEV}, derived directly from the UAV query, and the \textbf{Refined BEV}, which integrates geometric corrections from retrieved candidates. While the Initial BEV already provides a competitive baseline despite inherent translational and heading uncertainties, the Refined BEV effectively resolves these geometric offsets, yielding a substantial boost in retrieval performance. This confirms that our unified pipeline successfully transforms coarse initial estimates into precise geometric anchors.

\textbf{Investigation on IoU-based Retrieval Metric.} We analyze the retrieval of spatially overlapping neighboring tiles by varying the Intersection-over-Union (IoU) threshold $\tau$ in our $R_{\text{IoU}}@K$ metric. As shown in Table \ref{tab:iou_sensitivity}, relaxing the strict binary matching criterion to include overlapping candidates yields a consistent improvement in R@1 accuracy. This demonstrates that our method effectively retrieves neighboring tiles which, while distinct from the ground truth ID, share substantial visual content with the UAV query. At a threshold of $\tau=0.2$, the high recall rate confirms the retrieval of these geometrically relevant ``semi-positive'' samples. Crucially, we argue that these overlapping tiles serve as valid geometric anchors rather than false positives. By providing essential context, they enable the subsequent fine-grained pose estimation module to accurately recover the absolute UAV position, proving that effective geo-localization need not rely solely on rigid, non-overlapping tile identification.

\subsection{BEV View Alignment Results}

\textbf{BEV Refinement Iteration Analysis.} We quantify the impact of iterative BEV refinement on retrieval performance in Table \ref{tab:bev-refinement}. The initial BEV representation, synthesized exclusively from the UAV query, inherently suffers from geometric uncertainties---primarily translational shifts and heading misalignments relative to the canonical satellite grid. These discrepancies constrain the initial retrieval accuracy. However, by assimilating the retrieved satellite candidates into the scene reconstruction pipeline, the BEV representation undergoes iterative refinement. We observe that the primary geometric offsets are significantly rectified in the first iteration, yielding a substantial improvement in R@1. This validates the role of retrieved candidates as effective geometric priors. While subsequent iterations continue to refine the scene, the marginal performance gains diminish. Empirically, we find that two iterations strike an optimal balance between retrieval accuracy and computational efficiency.

\begin{figure}
    \centering
    \includegraphics[width=1.0\linewidth]{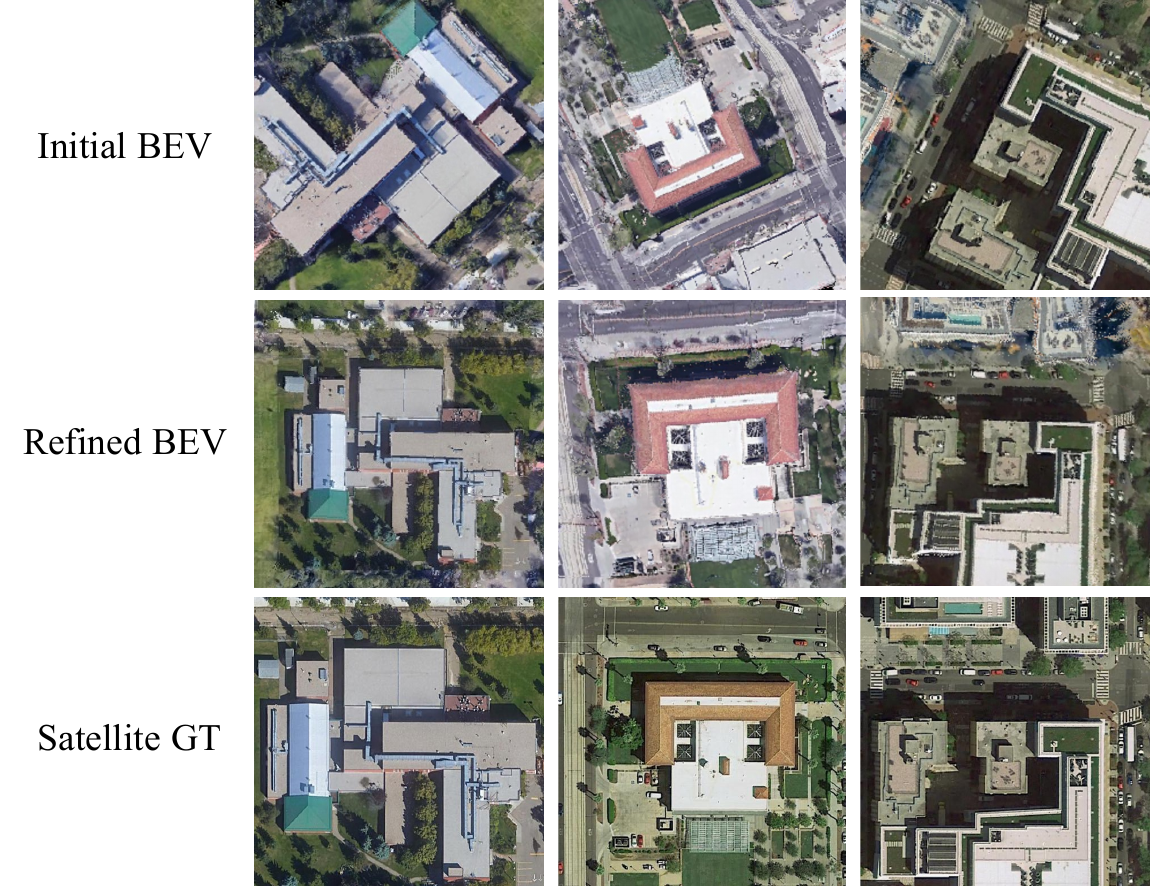}
    \caption{\textbf{Visualization of the BEV Refinement.}}
    \label{fig:bev-refinement}
    %\vspace{\fixedvskip}
\end{figure}

\textbf{Visualization of BEV Refinement.}
Qualitative analysis of the BEV refinement process is presented in Figure \ref{fig:bev-refinement}. The process initiates with a baseline BEV rendering derived solely from the UAV query via ground plane estimation. Constrained by the unknown positive satellite tile, this initial reconstruction often exhibits incomplete peripheral structures and geometric shift. Upon retrieving top-ranked satellite candidates, the system employs the \textit{Satellite-wise Attention Block} to spatially register these references against the initial scene. The retrieved satellite imagery supplements the reconstruction with complementary orthogonal context, effectively rectifying geometric distortions. The refined BEV output demonstrates precise spatial alignment with the ground truth satellite map, confirming that our unified framework successfully fuses cross-view information to mitigate the domain gap effectively.

\renewcommand{\arraystretch}{1.3}
\begin{table}
	\centering
	\caption{\textbf{Retrieval Results of BEV View Refinement.}}
	\begin{tabular}{c | c c c c }
        \toprule
        Refine Time & R@1 $\uparrow$  & R@5 $\uparrow$  & R@10 $\uparrow$  & AP $\uparrow$ \\
        \cline{1-5}
        0 (Initial BEV)	&58.8	&78.7	&84.2	&63.2 \\
        1	                &66.8	&84.0	&87.9	&70.6 \\
        \rowcolor{gray!20}  2	                &79.0	&88.0	&89.3	&81.1 \\
        3	                &80.5	&89.2 	&89.2	&82.2  \\
        \bottomrule
        \end{tabular}
	\label{tab:bev-refinement}
\end{table}

\subsection{3-DoF UAV Position Results}

\begin{figure}
    \centering
    \includegraphics[width=1.0\linewidth]{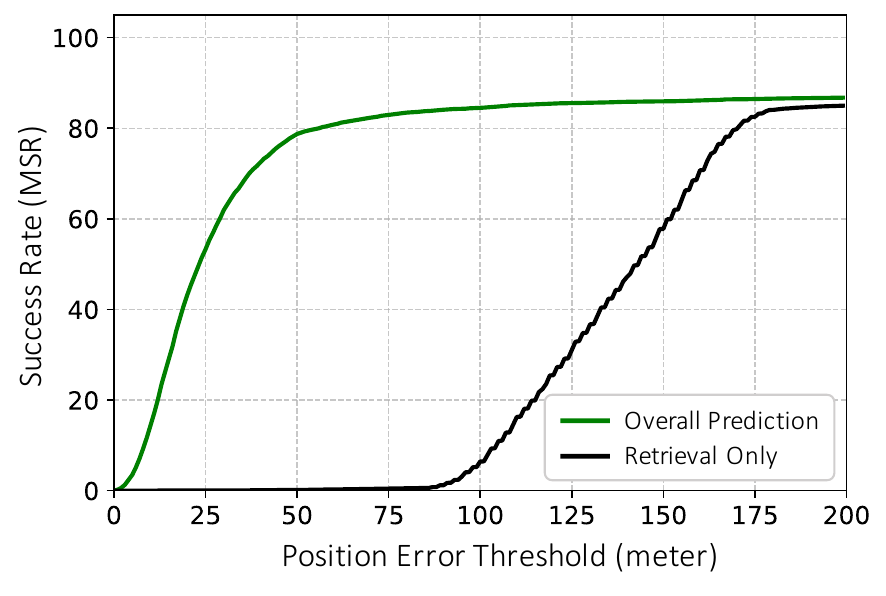}
    \caption{\textbf{Curve of Meter-level Position Success Rate.} Meter-level Success Rate (MSR@X) is the percentage of queries localized within the X-meter threshold.} 
    \label{fig:pose_curve}
\end{figure}

\renewcommand{\arraystretch}{1.3}
\begin{table}
	\centering
	\caption{\textbf{Result of Relative Pose Evaluation.} We report the 3-DoF Relative Pose Error (RPE) of UAV frames.}
    % \resizebox{\linewidth}{!}{
	\begin{tabular}{ c|c c c c}
        \toprule
        View Num. & $\hat E_{\text{RMSE}} \downarrow$ & $\hat E_{\text{median}} \downarrow$ & $\Delta \hat \theta_{\text{RMSE}} \downarrow$ & $\Delta \hat \theta_{\text{median}} \downarrow$ \\
        \cline{1-5}
        4   & 4.36  & 2.71	& 1.84	& 1.24\\
        12  & 4.65	& 2.85	& 1.96	& 1.03\\
        18  & 6.93	& 2.91	& 3.37	& 1.07\\
        \bottomrule
        \end{tabular}
	\label{tab:relative_pose_err}
\end{table}

\textbf{Global Pose Estimation.} We first evaluate the global pose estimation accuracy on geo-localization. The model ingests a sequence of UAV images and outputs the absolute position for every frame. To quantify performance, we measure the error between the predicted pose and the ground-truth trajectory. We compare our full geometry-aware model against a baseline that relies solely on the retrieval stage, visualizing the results via the meter-level position threshold curve shown in Figure \ref{fig:pose_curve}.

In conventional retrieval-based methods, the location of the UAV's position is naively assigned from the best-matching satellite tile. This inherently limits localization accuracy to the resolution of the satellite tiles. Consequently, as the success threshold drops below the tile size (e.g., $<$ 200 meters), the accuracy of the retrieval-only baseline degrades rapidly, even if the retrieval recall is high. In contrast, our proposed unified pipeline does not stop at retrieval; it further predicts the precise position within the retrieved tile. The results demonstrate a significant accuracy improvement, particularly at strict thresholds (e.g., 50 meters), validating that our pipeline successfully bridges the gap between tile-level coarse localization and meter-level fine-grained pose estimation.

\textbf{Relative Pose Performance.} We further assess the robustness of our internal geometric representation by evaluating the Relative Pose Error (RPE) between UAV frames. Table \ref{tab:relative_pose_err} summarizes the relative pose error $\hat E$ and heading error $\Delta \hat \theta$ across different view counts. The results indicate that the overall error remains stable regardless of view variance, underscoring the strong capability of the VGGT backbone to maintain consistent geometric reasoning. As the number of views increases, the heading error decreases, suggesting that broader visual coverage helps constrain orientation.  It shows that the system maintains high relative consistency, which is crucial for stable trajectory reconstruction.

\subsection{Ablation Study}

\renewcommand{\arraystretch}{1.3}
\begin{table}
	\centering
	\caption{\textbf{Ablation Study of the Proposed Modules.} (a) Satellite-wise Attention (b) Geometric Validation (c) Feature Validation .}
    % \resizebox{\linewidth}{!}{
	\begin{tabular}{c c c|c c c c}
        \toprule
        (a) & (b) & (c) & R@1$ \uparrow$ & $E_{\text{median}} \downarrow$ & $\Delta \theta_{\text{mean}} \downarrow$ & $\Delta \theta_{\text{median}} \downarrow$ \\
        \hline
        & & &67.2 &60.1	&22.9	&5.2 \\
        \checkmark & & &67.7	& 57.1	& 17.3	& 4.3 \\
        & &  \checkmark&71.6	 &42.1	 &19.3   &4.2 \\
        \checkmark &  \checkmark& & 72.3	& 37.8	& 20.7	&4.8\\
        \rowcolor{gray!20} \checkmark &  & \checkmark & \textbf{79.0}	 &\textbf{26.2}	  &\textbf{12.2} & \textbf{4.1}\\
        \bottomrule
        \end{tabular}
	\label{tab:ablation_modules}
\end{table}

\renewcommand{\arraystretch}{1.3}
\begin{table}
	\centering
	\caption{\textbf{Impact of the Number of Input Images.}}
    \tablestyle{6pt}{1.25}
    % \begin{threeparttable}
    \begin{tabular}{c | c c c c}  
    	\toprule
        View Number & R@1 $\uparrow$ & R@5 $\uparrow$ & R@10 $\uparrow$ & AP $\uparrow$ \\
        \cline{1-5}
        4	&64.9	&79.0	&82.9	&68.2 \\
        \rowcolor{gray!20} 12	&\textbf{79.0} 	&\textbf{88.0}	&89.3 	&81.1 \\
        18	&77.8	&86.4	&\textbf{90.6}	&\textbf{83.1} \\
        \bottomrule
    \end{tabular}
    \label{tab:view_number}
\end{table}
To empirically validate the contributions of our proposed modules, we conducted a comprehensive ablation study on the University-Pose dataset. We systematically investigate the impact of the \textit{(a) Satellite-wise Attention Block} and evaluate two distinct candidate verification strategies: \textit{(c) Geometric Validation} and \textit{(c) Feature Validation}. The quantitative performance, measured by Refined BEV Recall (R@1), absolute global positioning error ($E$), and heading error ($\Delta \theta$), is detailed in Table \ref{tab:ablation_modules}.

The baseline configuration (Row 1) employs a vanilla global attention mechanism without candidate verification, resulting in a suboptimal mean positioning error. We attribute this degradation to hypothesis interference where independent satellite candidates act as competing noise sources within the shared attention field, confounding the model's ability to isolate the correct geometric context. Furthermore, the absence of a verification mechanism renders the model unable to filter out false positive retrieval targets, thereby degrading localization accuracy.

\textbf{Impact of Satellite-wise Attention.} The integration of the \textit{(a) Satellite-wise Attention Block} structurally isolates the interaction between each satellite candidate and the UAV query. This modification yields a substantial performance gain in pose estimation. This improvement confirms that suppressing inter-satellite interference allows the model to extract distinct, discriminative geometric features for each hypothesis, thereby refining the precision of pose regression.

\textbf{Comparison of Validation Strategies.} We further evaluate two strategies for selecting the optimal satellite reference from the top-$k$ candidates:
\begin{itemize}
\item \textit{Geometric Validation (b)}: Candidates are selected based on point cloud confidence and pose consistency. While this approach effectively filters outliers with poor geometric convergence, it struggles to distinguish between spatially similar but distinct locations and relies heavily on sensitive, manually tuned thresholds.
\item \textit{Feature Validation (c)}: Prioritizing candidates with the highest implicit feature similarity to the UAV query results in superior error reduction. Txhis finding underscores that implicit feature consistency offers a more robust signal for place recognition than geometric statistics alone, as it effectively discriminates against semantically mismatched distractors that may otherwise satisfy loose geometric constraints.
\end{itemize}

\begin{figure}
    \centering
    \includegraphics[width=1\linewidth]{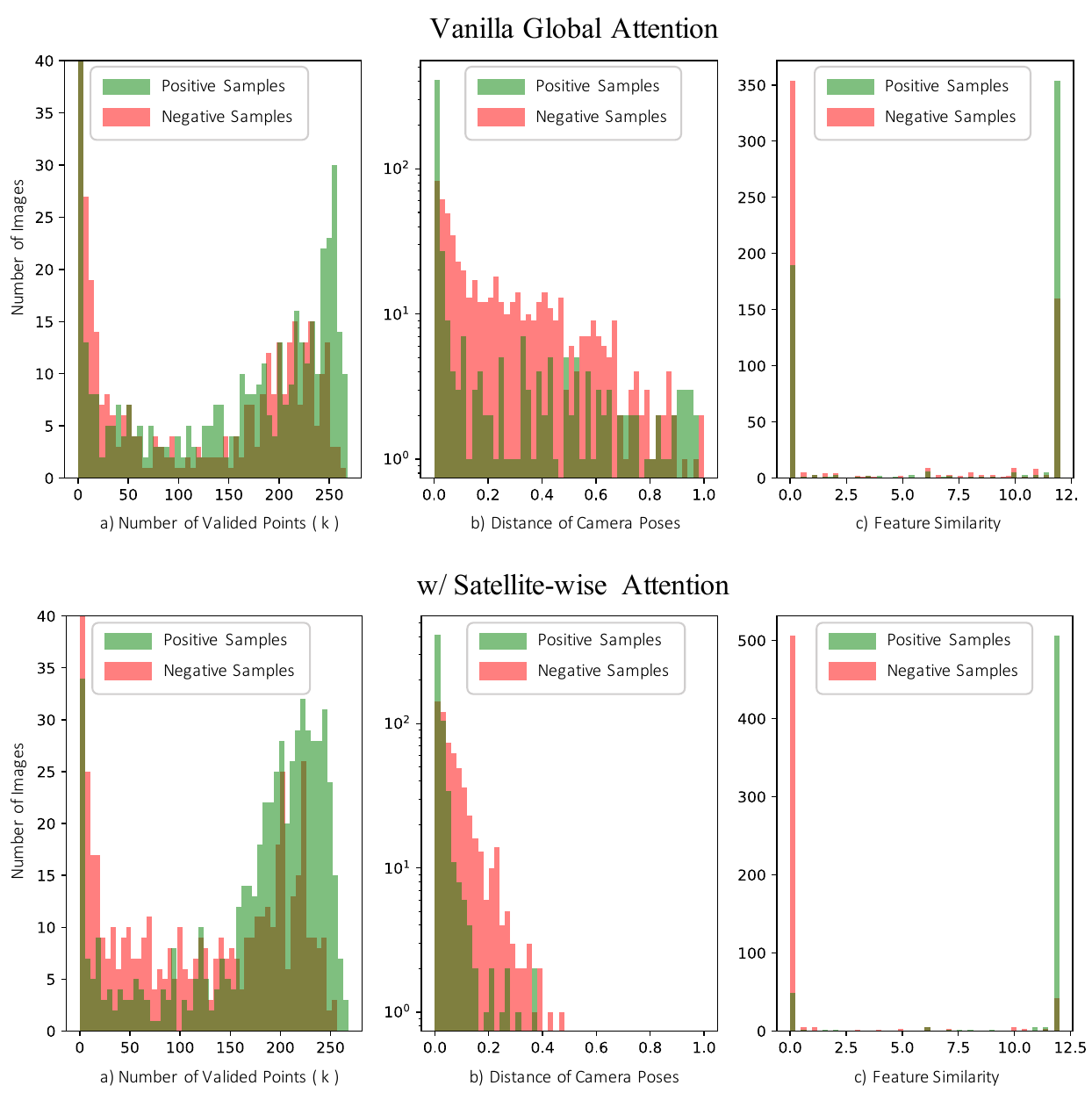}
    \caption{\textbf{Overlapped Satellite Images Distribution Comparison.}} 
    \label{fig:sat-stastic}
\end{figure}

\subsection{Impact of Input Sequence}
We investigate the influence of the UAV input sequence length on the overall geo-localization performance, as detailed in Table \ref{tab:view_number}. The efficacy of the 3D perspective model is inherently dependent on the availability of multi-view observations to accurately reconstruct the scene geometry and align it with the satellite perspective. With sparse input sequences (e.g., fewer than 5 frames), we observe a marked degradation in retrieval performance, dropping to approximately 64\%. This decline is attributed to the insufficient geometric cues required to constrain the 3D structure, leading to distorted BEV synthesis and poor cross-view alignment. As the sequence length increases, performance improves substantially, showing a robust balance of geometric coverage and feature redundancy. Extending the sequence beyond 18 views yields performance that fluctuates slightly due to the accumulation of feature noise, while the computational cost escalates linearly.

\subsection{Feature Distribution Analysis}

In Figure \ref{fig:sat-stastic}, we analyze the distribution of positive (green) and negative (red) satellite candidates across three key metrics: a) the number of predicted 3D points exceeding a 20\% confidence threshold, b) the spatial distance between the predicted camera pose and the estimated BEV center, and c) the cosine feature similarity between candidate and UAV tokens. We compare these distributions under both the baseline vanilla global attention and our proposed Satellite-wise Attention.

\textbf{Effectiveness of Satellite-wise Attention.} A column-wise comparison reveals that introducing Satellite-wise Attention consistently enlarges the separation between positive and negative candidates across all three metrics. This effect is particularly evident in metric (b), which directly reflects geometric consistency. Under vanilla global attention, predicted camera poses exhibit large variance and frequently drift away from the BEV center, indicating that irrelevant satellite candidates interfere with the shared attention context. In contrast, Satellite-wise Attention enforces independent interactions between each satellite candidate and the UAV scene, effectively isolating mutually exclusive hypotheses. As a result, valid candidates concentrate tightly around the BEV center, forming a compact and coherent geometric distribution that is more amenable to accurate pose regression.

\textbf{Superiority of Feature Similarity Verification.} A row-wise comparison further highlights the relative effectiveness of different verification signals. Although explicit geometric cues (a) the number of valid 3D points and (b) pose distance exhibit mild distributional shifts between positive and negative samples, substantial overlap remains, making it difficult to define reliable handcrafted thresholds. In contrast, the feature similarity metric in (c) achieves a markedly clearer separation, which is further amplified by Satellite-wise Attention. This observation indicates that learned feature similarity provides a more discriminative and robust criterion for candidate verification than standalone geometric heuristics. Moreover, applying a softmax operation sharpens the similarity distribution into a sparse and peaked profile, suggesting that the model can reliably identify the true satellite candidate within a single forward pass, without complex heuristic filtering.

\subsection{Qualitative Results}
\begin{figure*}[t]
    \centering
    \includegraphics[width=1.0\linewidth]{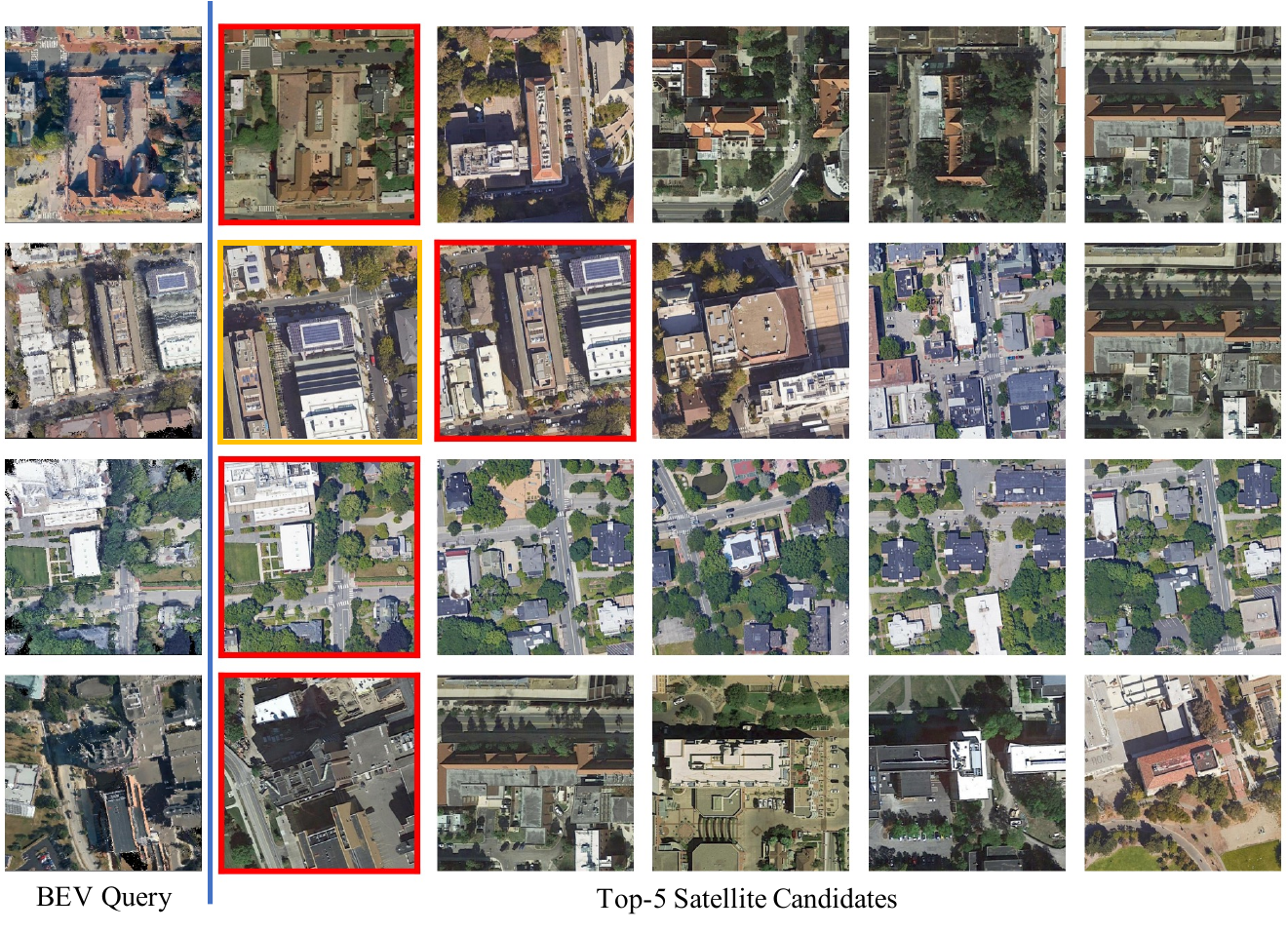}
    \caption{\textbf{Visualization of Retrieval Results.}}
    \label{fig:vis_retrieval}
    %\vspace{\fixedvskip}
\end{figure*}

We present qualitative visualizations to substantiate the efficacy of our unified retrieval and pose estimation pipeline. Figure \ref{fig:vis_retrieval} showcases the top-5 retrieved satellite candidates for challenging UAV query sequences. The first column depicts the synthesized BEV query derived from the oblique UAV inputs. Within the retrieval rankings, the ground truth satellite tile is delineated by a red bounding box, while neighboring tiles sharing significant spatial overlap are highlighted in yellow. The frequent high ranking of these overlapping candidates indicates that our model effectively learns underlying geometric consistency rather than relying solely on transient texture patterns. This observation empirically validates the utility of our IoU-based retrieval metric, which credits geometrically relevant matches.

\begin{figure}
    \centering
    \includegraphics[width=1.0\linewidth]{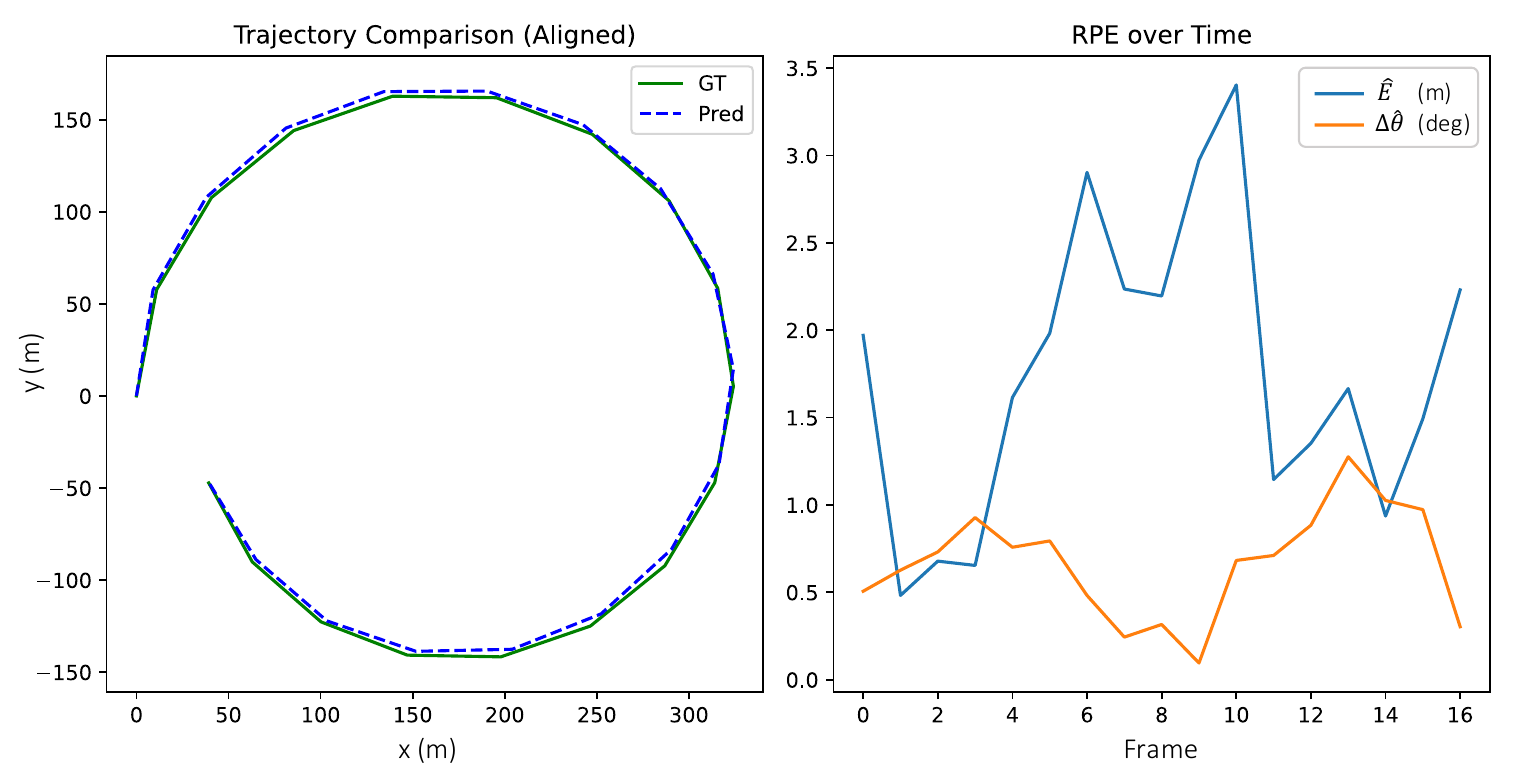}
    \caption{\textbf{Visualization of the UAV Trajectory.}}
    \label{fig:vis_pose}
    %\vspace{\fixedvskip}
\end{figure}

Furthermore, Figure \ref{fig:vis_pose} illustrates the fine-grained localization performance. We observe a precise alignment between the predicted UAV trajectories and the ground truth poses, maintaining accuracy even in complex urban environments characterized by repetitive structures. This visual correspondence corroborates the low quantitative error metrics reported in Table \ref{tab:relative_pose_err}, affirming the system's capability to deliver robust and consistent pose estimation over extended flight durations.

\section{Discussion}
\label{sec:discussion}

To further advance the field of unified cross-view geo-localization, this section focuses on addressing the following three critical questions:

\textbf{How to train this model with end-to-end learning strategy?} While our current framework leverages the robust zero-shot capabilities of pre-trained foundation models like DINOv2 and VGGT to bridge the domain gap, we recognize that these backbones are primarily optimized for general object classification or generic scene reconstruction rather than the specific nuances of cross-view geo-localization. Consequently, a domain gap persists between the natural images used for pre-training and the orthorectified satellite imagery encountered in our task. To address this, future research of this work should prioritize an end-to-end training strategy. The inherent differentiability of the VGGT rendering process offers a unique opportunity to propagate geometric alignment errors directly back to the feature extraction layers. By formulating a unified objective function that couples retrieval loss with pose regression accuracy, the network could be fine-tuned to learn feature descriptors that are explicitly invariant to the facade-to-roof transformation. Such a strategy would not only adapt the model to the specific texture distributions of satellite orthophotos but also allow the geometric reconstruction module to prioritize structural elements critical for geo-localization---such as road topologies and building footprints---over transient dynamic objects, thereby maximizing the efficacy of the proposed 3D scene alignment paradigm.

\textbf{How to work with single-view unified perception?} A principal limitation of the proposed model is its reliance on multi-view image sequences to solve the reconstruction problem, which imposes specific flight trajectory requirements and latency overheads. Our experiments indicate that reducing the number of input views degrades the geometric fidelity of the reconstructed scene, subsequently impairing cross-view alignment. However, the requirement for multi-view parallax is not an insurmountable barrier for the unified model architecture. Recent advancements in monocular depth estimation and generative diffusion models demonstrate that strong geometric priors can be learned from large-scale data, enabling the recovery of 3D structure from single images. We posit that the future of this unified framework lies in integrating such monocular depth priors directly into the scene representation module. By hallucinating the occluded geometry and recovering scale through learned semantic cues, the model could theoretically synthesize a reliable BEV representation from a single UAV frame. This evolution would decouple the system from the necessity of extended image sequences, extending the applicability of our geometry-aware alignment to instantaneous, snapshot-based localization scenarios where flight motion is restricted, or immediate positioning is critical.

\textbf{How to achieve end-to-end implicit feature alignment instead of explicit BEV construction?} Currently, our pipeline necessitates the explicit rendering of a virtual BEV image to bridge the viewpoint discrepancy, a process that inherently incurs information loss through quantization, rasterization, and the projection of high-dimensional semantics into a lower-dimensional RGB or feature space. While effective, this intermediate representation acts as a bottleneck that filters out potentially discriminative uncertainty information and subtle feature correlations. We argue that the next logical step in ``3D-aware'' localization is to transcend explicit visual reconstruction in favor of implicit feature-level alignment. Future work could explore the construction of Neural Feature Fields, where the retrieval and localization processes occur directly via cross-attention mechanisms between satellite patch tokens and the 3D voxel tokens of the reconstructed scene. By skipping the generation of a 2D BEV image, the system could maintain the full integrity of the high-dimensional feature manifold, allowing the satellite queries to attend directly to the underlying 3D geometry. This would represent a shift towards a fully differentiable, feature-centric alignment pipeline that preserves rich semantic context and geometric probability distributions, further pushing the boundaries of accuracy in complex, large-scale environments.

% conclusion
\section{Conclusion}
\label{sec:conclusion}
In this work, we have presented a unified framework for UAV cross-view geo-localization that overcomes the limitations of conventional decoupled retrieval and pose estimation pipelines. Instead of treating coarse place recognition and fine-grained localization as independent stages, our approach establishes a shared 3D geometric representation in which both tasks are jointly supported during inference. By leveraging geometry-grounded visual perception, we bridge the extreme viewpoint discrepancy between oblique UAV imagery and orthogonal satellite views through an intermediate BEV-aligned representation. This geometry-consistent representation enables robust feature alignment and stable pose reasoning without relying on task-specific training. Importantly, we further address the challenge of multi-candidate ambiguity by introducing the Satellite-wise Attention Block, which enforces hypothesis isolation and prevents interference among mutually exclusive satellite hypotheses, leading to more coherent geometric reasoning. Extensive experiments on the recalibrated University-1652 and SUES-200 benchmarks demonstrate that the proposed framework achieves strong zero-shot geo-localization performance, delivering reliable meter-level pose estimation under challenging cross-view conditions.

% \section*{Acknowledgements}
% The research was partially supported by the National Natural Science Foundation of China (NSFC) under Grants 62271355. The numerical calculations were conducted on the supercomputing system in the Supercomputing Center, Wuhan University. 

% \printcredits

%\bibliographystyle{ieee_fullname}
\bibliography{Jinwang-Papers, chang, ref}
%\printbibliography
% \end{sloppypar}
\end{document}